\documentclass[conference]{IEEEtran}
\IEEEoverridecommandlockouts
\usepackage{cite}
\usepackage{amsmath,amssymb,amsfonts}
\usepackage{algorithmic}
\usepackage{graphicx}
\usepackage{textcomp} 
\usepackage{url}
\usepackage{cleveref} 
\usepackage{comment}
\usepackage{xcolor}
\usepackage{colortbl}
\def\BibTeX{{\rm B\kern-.05em{\sc i\kern-.025em b}\kern-.08em
    T\kern-.1667em\lower.7ex\hbox{E}\kern-.125emX}}
\begin{document}

\newcommand{\mcz}[1]{\multicolumn{2}{|c|}{#1}}
\newcommand{\mcd}[1]{\multicolumn{3}{|c|}{#1}}

\title{WSESeg: Introducing a Dataset for the Segmentation of Winter Sports Equipment with a Baseline for Interactive Segmentation}

\author{\IEEEauthorblockN{Robin Schön}
\IEEEauthorblockA{\textit{Fakultät für Angewandte Informatik} \\
\textit{University of Augsburg}\\
Augsburg, Germany \\
robin.schoen@uni-a.de}
\and
\IEEEauthorblockN{Daniel Kienzle}
\IEEEauthorblockA{\textit{Fakultät für Angewandte Informatik} \\
\textit{University of Augsburg}\\
Augsburg, Germany \\
daniel.kienzle@uni-a.de}
\and
\IEEEauthorblockN{Rainer Lienhart}
\IEEEauthorblockA{\textit{Fakultät für Angewandte Informatik} \\
\textit{University of Augsburg}\\
Augsburg, Germany \\
rainer.lienhart@uni-a.de}
}

\maketitle

\begin{abstract}
In this paper we introduce a new dataset containing instance segmentation masks for ten different categories of winter sports equipment, called WSESeg (Winter Sports Equipment Segmentation) \footnote{Available at \url{https://github.com/Schorob/wseseg} .}. Furthermore, we carry out interactive segmentation experiments on said dataset to explore possibilities for efficient further labeling. 
The SAM and HQ-SAM models are conceptualized as foundation models for performing user guided segmentation. In order to measure their claimed generalization capability we evaluate them on WSESeg. Since interactive segmentation offers the benefit of creating easily exploitable ground truth data during test-time, we are going to test various online adaptation methods for the purpose of exploring potentials for improvements without having to fine-tune the models explicitly. Our experiments show that our adaptation methods drastically reduce the Failure Rate (FR) and Number of Clicks (NoC) metrics, which generally leads faster to better interactive segmentation results.  
\end{abstract}

\begin{IEEEkeywords}
interactive segmentation, instance segmentation, sports, winter sports, ski, snowboards, winter sports equipment, dataset
\end{IEEEkeywords}

\section{Introduction}
There is an interest in the exact analysis of the pose and limb positions of a human being depicted in an image or video. The corresponding computer vision tasks are human pose estimation \cite{andriluka14cvpr} and body part segmentation \cite{Li2017MultipleHumanPI, gonginstance}, where progress mainly benefits the analysis of sports-related image and video data. 
In some cases deep learning applications can be applied for the post hoc analysis of the athletic performance with the aim of finding room for improvement. 
For this purpose it is necessary to detect the exact position of each limb of the body. 

The authors of \cite{ludwig2023all, ludwig2023detecting} leverage segmentation masks for body parts of human athletes as well as equipment to train a network capable of localizing any desired keypoint on the human and their equipment. There is a wide availability of datasets containing human poses in a skeletal form and body part segmentation masks for human limbs. However, data concerning the equipment used in the sports is relatively scarce, which lead to the authors resorting to pseudo labels. 
This insufficient availability of data particularly concerns the winter sports domain, which is why we created a dataset containing segmentation masks for various types of winter sports equipment worn or used by the respective athletes. 

Most available datasets containing segmentation masks only provide these types of annotations for general consumer images. Often, there is a lack of segmentation masks for rare applications that considerably differ from the domain of general consumer images. In these cases, there will be a need to annotate ground truth segmentation masks for the new purpose. However, directly annotating these masks generally constitutes a considerable amount of work. This is especially the case when the segmentation masks are annotated by the means of drawing polygons (as in the case of COCO \cite{lin2014microsoft}) and the annotator has to draw masks for fine structures. In order to alleviate this problem, there has been a considerable amount of research dedicated to the development of systems that are able to infer a segmentation mask from simple, quickly providable user guidance. In the most cases, this guidance amounts to clicks or scribbles on the foreground or background to indicate the position of the object. 

Such interactive segmentation systems should be applicable to a wide range of data. The authors of \cite{kirillov2023segment} present the \emph{Segment Anything Model (SAM)} as a foundation model that aims at generalizing to unseen domains or types of objects. 
The authors of \cite{sam_hq} present HQ-SAM, which is a slight modification of SAM. It is geared towards producing segmentation masks with a higher quality and endowing the model with the capacity to segment finer structures. For this purpose, HQ-SAM was additionally fine-tuned with the high-quality human annotated masks from HQSeg-44k. In contrast, SAM has only been trained on automatically generated masks. 
Despite the extensive training, foundation models such as SAM and HQ-SAM often face issues when exposed to domains that do not resemble their training data \cite{chen2023sam}. 
In order to rectify these limitations, we will explore the applicability of test-time adaptation (TTA) methods when adapting SAM and HQ-SAM to this winter sports dataset. Our experiments will show that there is a considerable room for improvement when applying TTA. 
Our contributions can be summarized as follows: 
\begin{itemize}
    \item We provide a dataset which contains high-quality instance masks for 10 types of winter sports equipment. We call this dataset WSESeg. \footnote{Dataset, benchmark and interactive segmentation baseline will be released on Github right before publication.} 
    \item We explore the usability of SAM and HQ-SAM for the usage as interactive segmentation systems on our winter sports dataset. We measure their performance in terms of the Failure Rate and Number of Clicks metrics.
    \item We compare various schemes for the test-time adaptation of interactive segmentation models to boost performance compared to the standard versions of SAM and HQ-SAM. 
\end{itemize}


\section{The Winter Sports Equipment Segmentation (WSESeg) Dataset} 
There has already been a considerable amount of literature and datasets discussing the importance of being able to segment the body parts, clothing and even certain types of equipment a person is wearing. 
The authors of \cite{Li2017MultipleHumanPI, gonginstance} have published datasets for the task of segmenting all parts of a human body, including clothing items. 
In addition to that, the usage of segmentation masks of sports equipment is discussed in  \cite{ludwig2023all, ludwig2023detecting, deepsportsradarv2}. 
The authors of \cite{sawahata2024instance} demonstrate a viable use case for segmentation masks of sports equipment, by segmenting swords used in fencing. 

With the aim of contributing to this line of research, we publish a novel dataset containing instance segmentation masks of winter sports equipment. The published dataset is called WSESeg (Winter Sports Equipment Segmentation).
The datasets contains ten classes. This corresponds to nine different types of object. Skis occur in the form of two classes: One class specifically contains skis in the context of ski jumping, while the other class contains skis used in other contexts. 

An overview over the classes as well as the amount of images and masks that can be found in each class are given in \Cref{tab:dataset}. 
Most of the images have been collected from Flickr, by automatically downloading all search results corresponding to a certain query (one query for each class). Afterwards the resulting set of images has been manually filtered to only contain images that actually display the desired type of object. 
The only exception is provided by the class for skis in the context of ski jumping, where the images originate from the first video of the YouTube Skijump Dataset \cite{ludwig2023detecting}. The chosen frames are exactly the ones indicated by the authors. 
In order to provide high quality annotations in the form of fine-grained masks, we used an existing interactive segmentation system \cite{liu2022simpleclick}.

In addition to this, any form of segmentation masks for rarely occurring objects constitutes a viable way of evaluating interactive segmentation systems, as the authors of \cite{kirillov2023segment} demonstrate. 
SAM \cite{kirillov2023segment} and HQ-SAM \cite{sam_hq} have been trained for the usage of general consumer images. More specific types of items, such as winter sports equipment in a scenery containing a considerable amount of snow, are rare in such image sets. 
We are going to evaluate the viability of these systems for the task of interactive segmentation on the rare domains provided by our WSESeg dataset. 
Furthermore, we are going to test various tactics geared towards improving the systems performance during usage time. These methods are described in \Cref{sec:method}. 
Another reason why our data may be challenging for a model lies in the great variety of object sizes. Bobsleighs constitute the class with the largest objects covering 13.05 \% of the area of the respective images, while the ski of skijumpers cover only 0.78 \% of the images. Although the COCO dataset (see \cite{lin2014microsoft}) also offers masks for skis, they only provide the mask annotations in the form of rather coarse polygons.

\begin{figure*} 
    \centering
    {\def\plotheighta{0.125\linewidth}
    \def\plotheightb{0.125\linewidth}
    \includegraphics[height=\plotheighta]{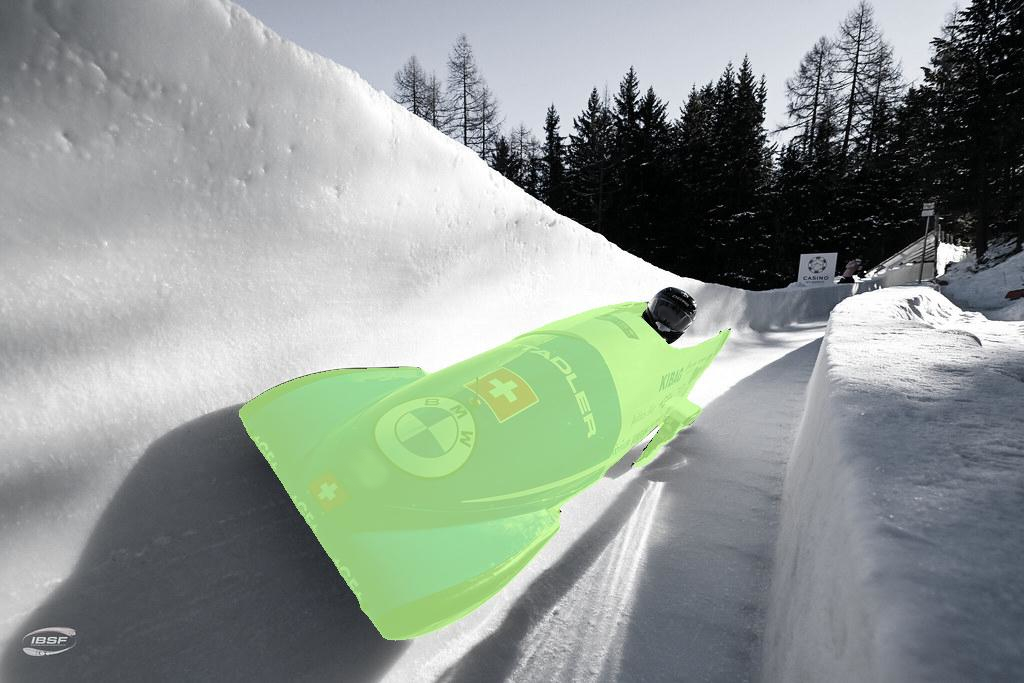} \hfill 
    \includegraphics[height=\plotheighta]{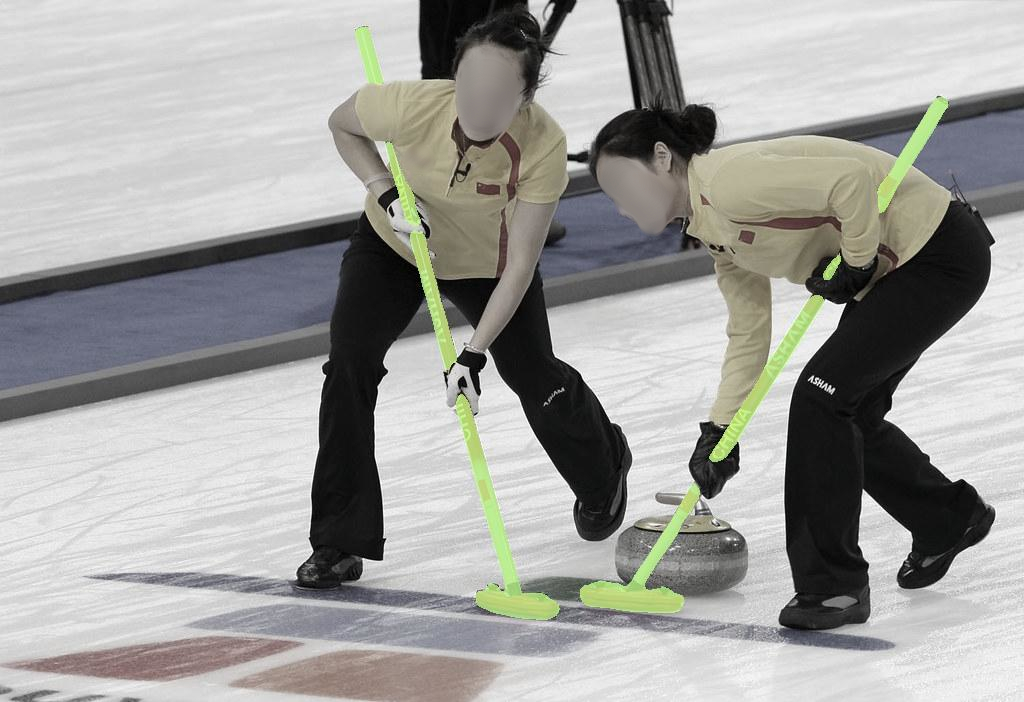} \hfill 
    \includegraphics[height=\plotheighta]{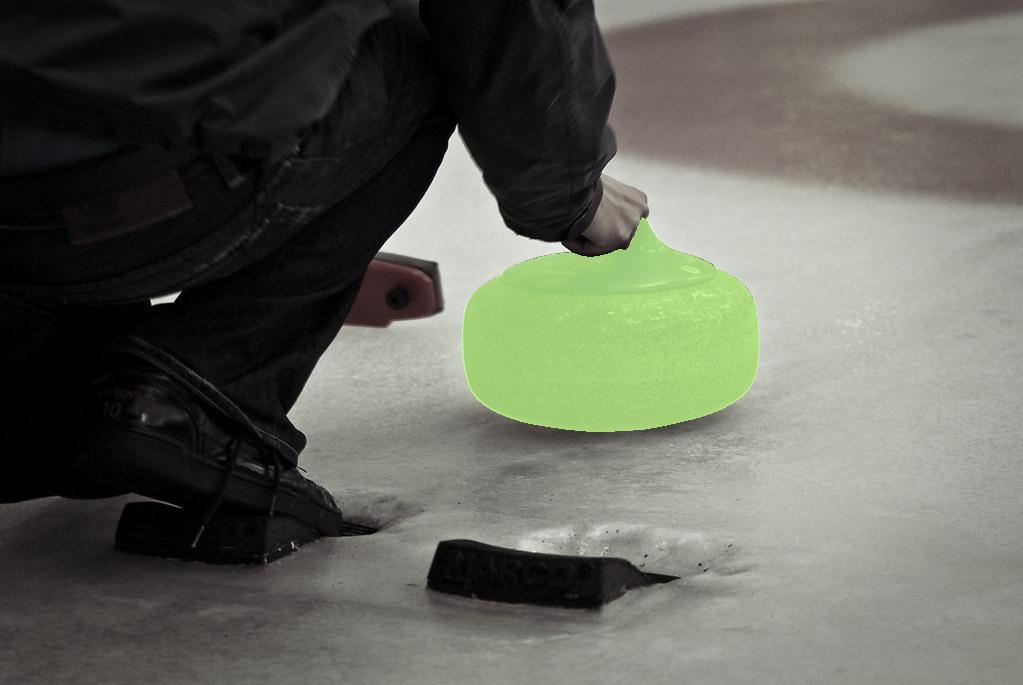} \hfill 
    \includegraphics[height=\plotheighta]{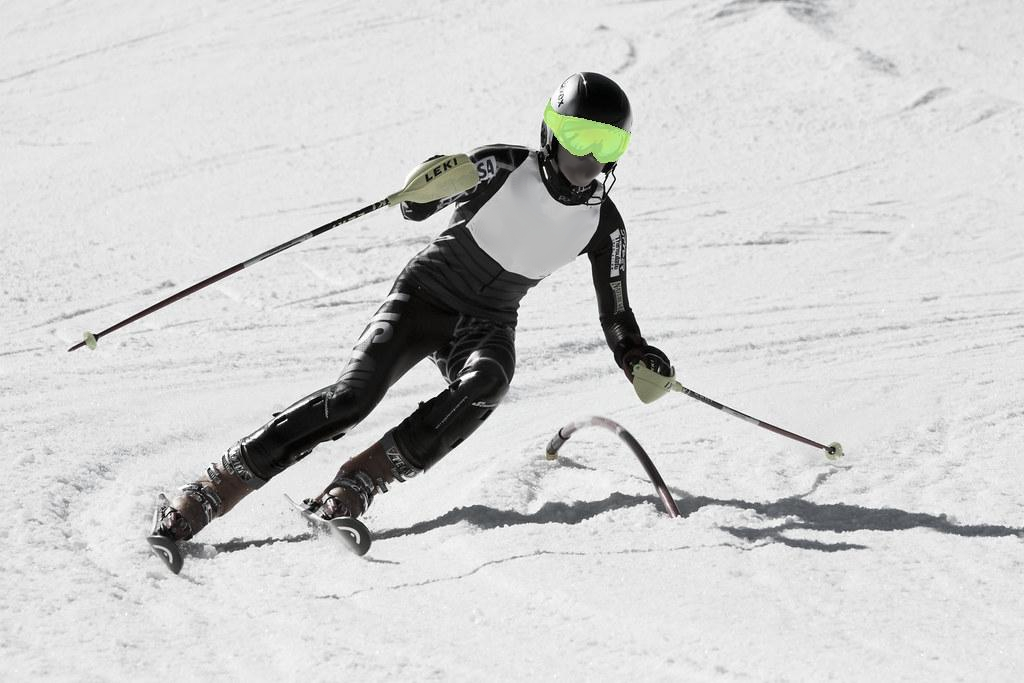} \hfill 
    \includegraphics[height=\plotheighta]{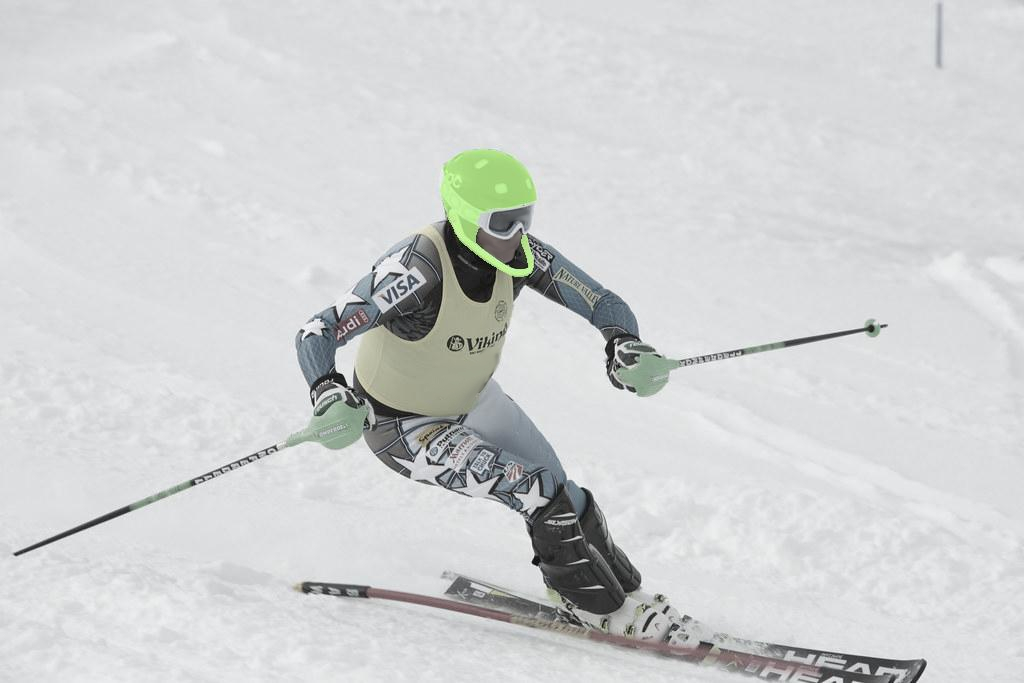} \\
    \vspace{3mm}
    \includegraphics[height=\plotheightb]{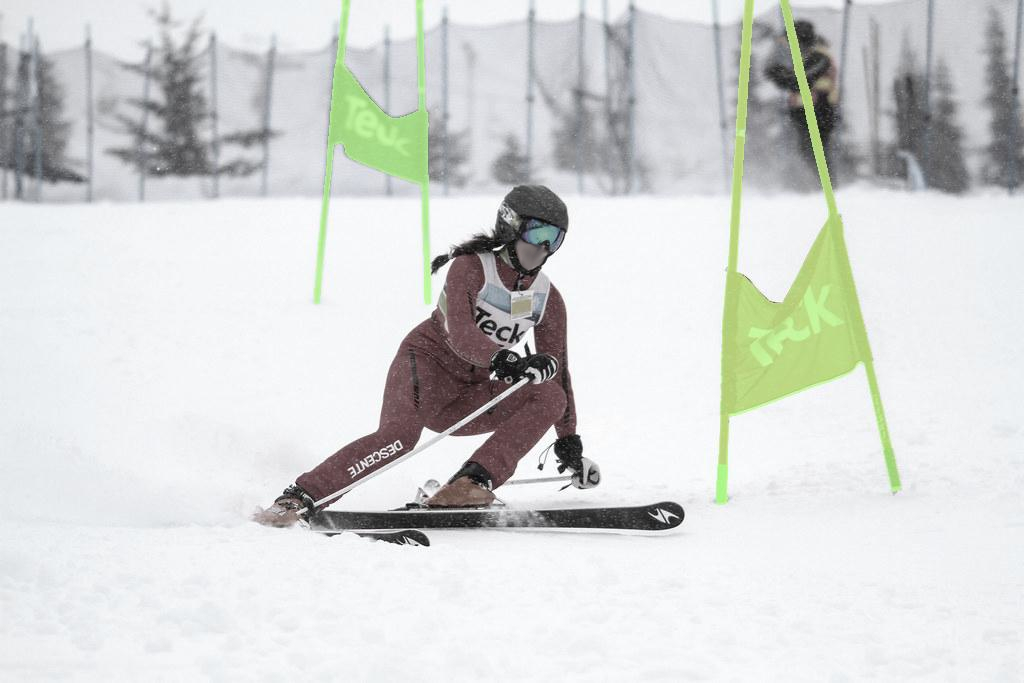} \hfill 
    \includegraphics[height=\plotheightb]{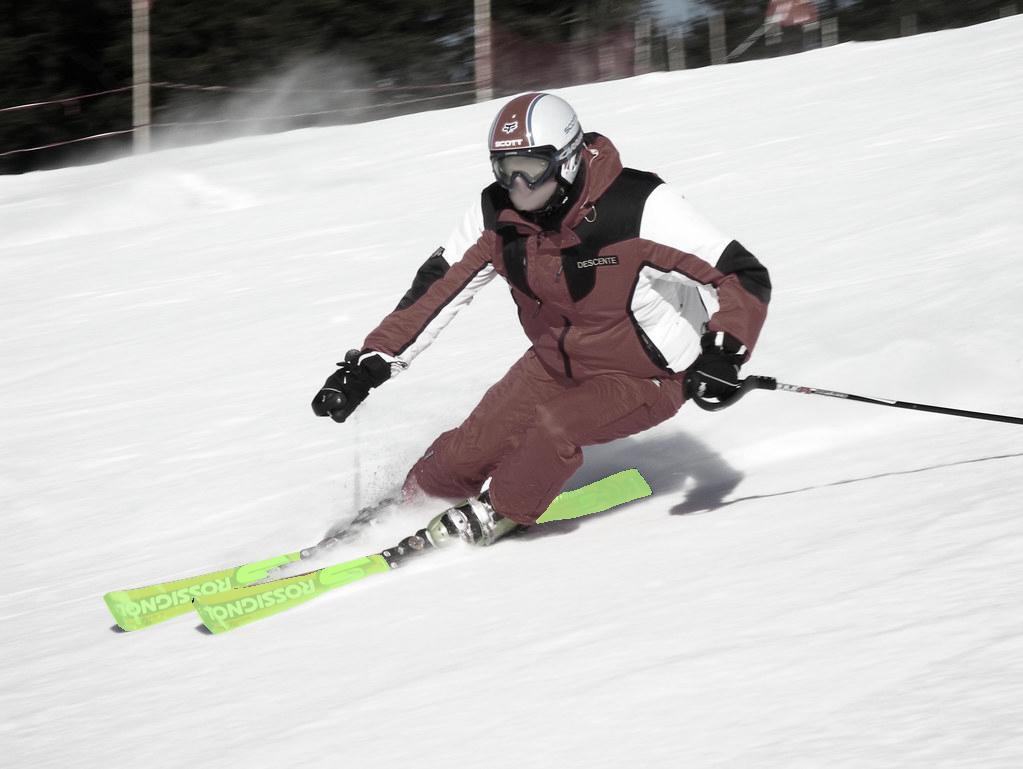} \hfill 
    \includegraphics[height=\plotheightb]{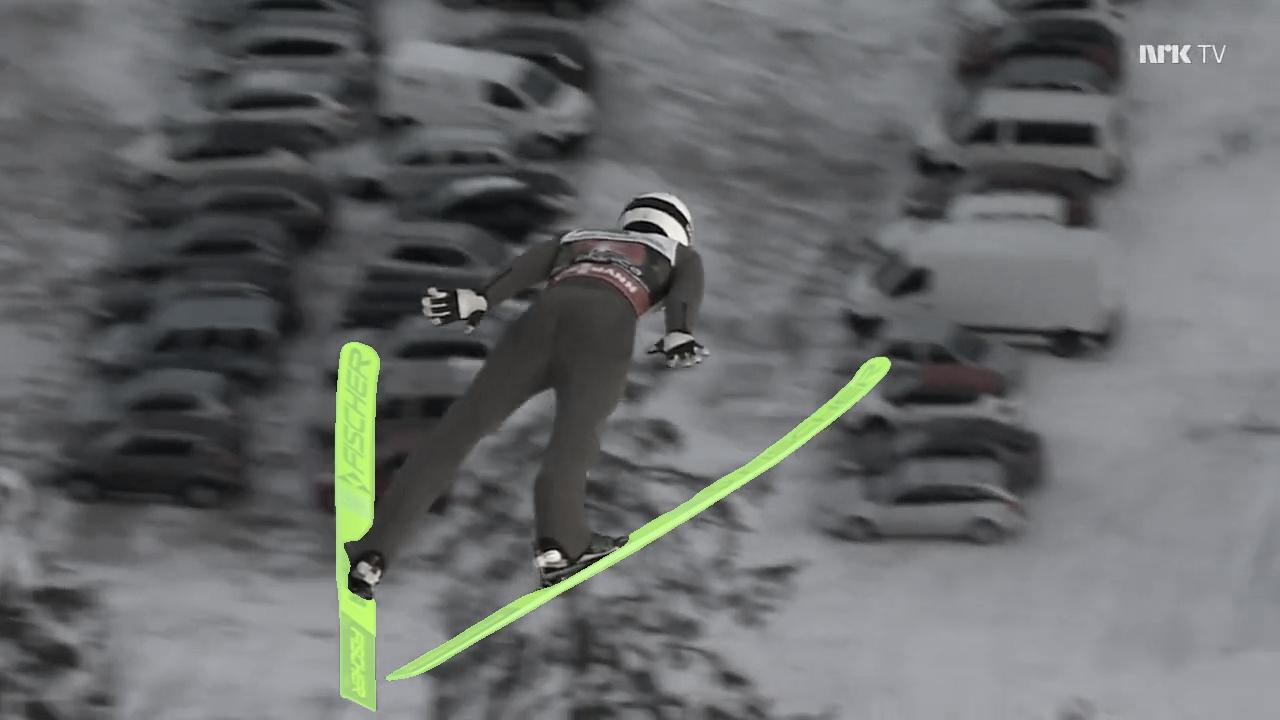} \hfill 
    \includegraphics[height=\plotheightb]{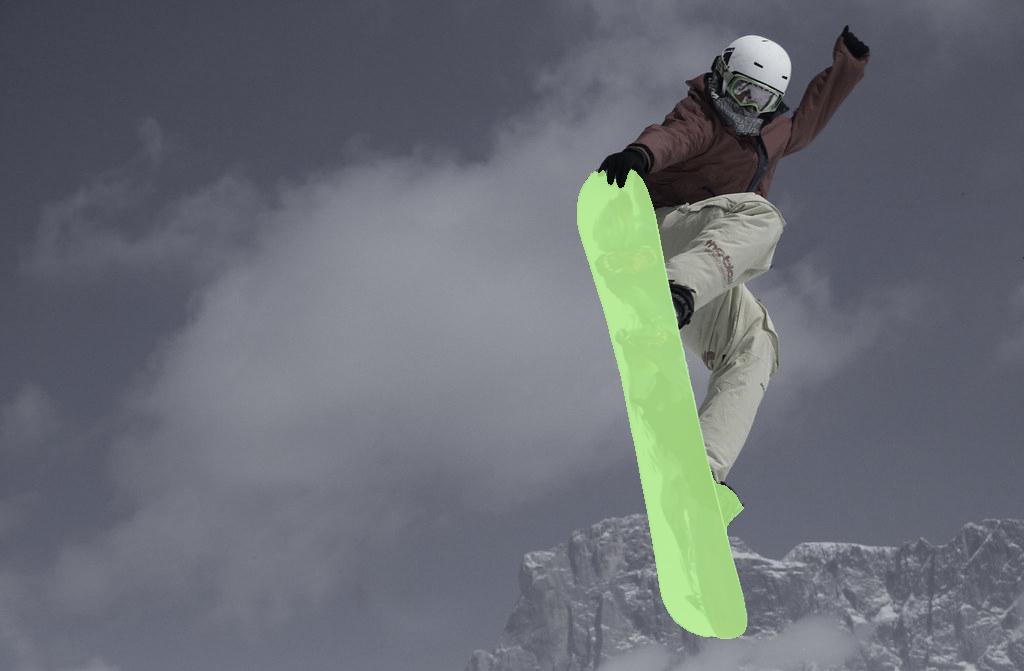} \hfill 
    \includegraphics[height=\plotheightb]{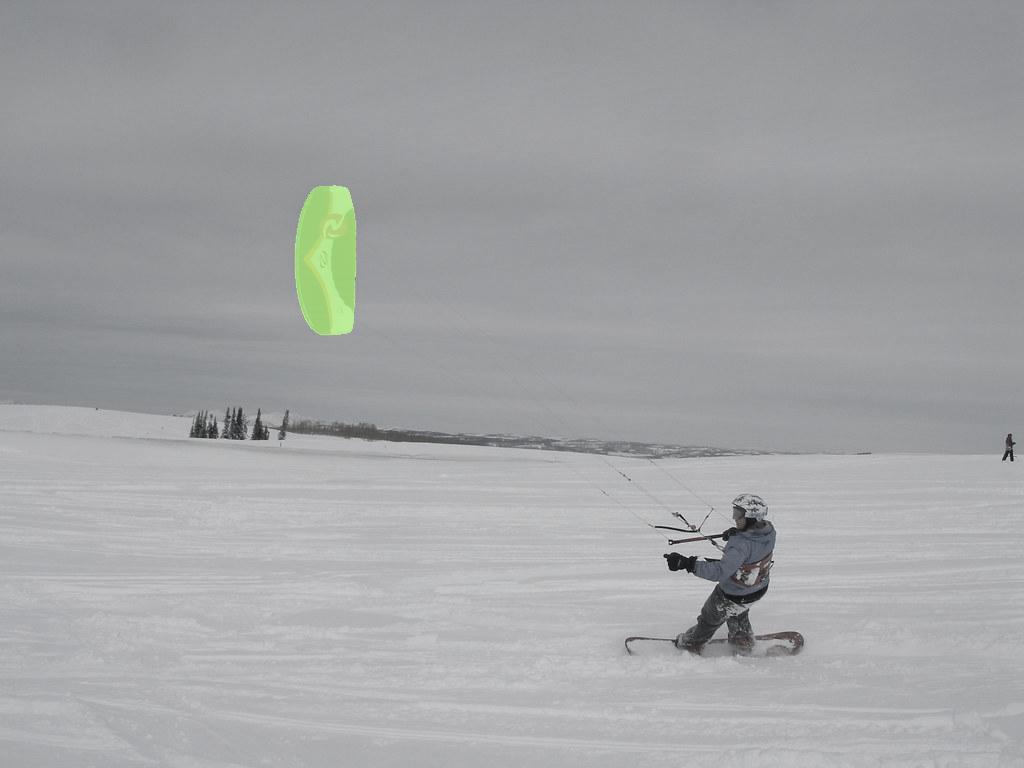}  
    \caption{Sample images from the WSESeg dataset. The object types being segmented are the following. Upper row: Bobsleigh, Curling Broom, Curling Stone, Ski Goggles, Ski Helmet. Lower Row: Slalom Gate Poles, Skis (Misc), Skis (Skijump), Snowboard, Snowkite. The saturation of the images has been decreased for better visibility.}
    \label{fig:wseseg_samples}}
\end{figure*}

A selection of sample images can be seen in Fig.\ref{fig:wseseg_samples}.

\begin{table}[]
    \centering
    \caption{The number of masks and images in each class of the novel WSESeg dataset. }
    \label{tab:dataset}
    \begin{tabular}{|c|c|c|} 
         \hline
         \textbf{Class Name} & \textbf{\# Masks} & \textbf{\# Images}\\
         \hline
         Ski (Jump) & 498 & 249  \\
         Ski (Misc) & 601 & 245 \\ 
         Bobsleighs & 620 & 572 \\
         Curling Brooms & 656 & 284 \\
         Curling Stones & 983 & 285 \\
         Ski Goggles & 599 & 501 \\
         Ski Helmets & 684 & 555 \\
         Slalom Gate Poles & 1034 & 507 \\
         Snowboards & 650 & 491 \\
         Snow Kites & 1127 & 532 \\
         \hline 
         \textbf{Total} & 7452 & 4221 \\
         \hline
    \end{tabular}
\end{table}

\section{Related Work} 
\subsection{Segmentation of Instances in Sports}
The authors of \cite{ghasemzadeh2021deepsportlab} propose a system for the tasks of player instance segmentation and ball localization. Their paper points out the importance of segmenting sports equipment.
In \cite{gao2023sparse} we find a player instance segmentation system for the DeepSportsRadar dataset. 
The authors of \cite{Cioppa_2019_CVPR_Workshops} develop a real time system for semantic segmentation. The players are viewed as a single mask. 
\cite{sawahata2024instance} provides us with another perspective of why the tracking of sports gear is important. Therein the authors present a system for tracking the sword during fencing. In addition to performing instance segmentation on the sword itself, the tip of the sword is tracked in the form of a keypoint.  

\subsection{Interactive Segmentation} 
Interactive segmentation deals with methods to segment objects in images with the help of repeated user interaction (see \cite{li2018interactive, chen2021conditional, chen2022focalclick}). In many cases this interaction takes the form of clicks provided by the user (see \cite{ritm2022,sofiiuk2020f} and \cite{jang2019interactive}). 
The authors of \cite{kirillov2023segment} present the segment anything model (SAM), which has been trained on SA-1B, a dataset containing 1.1B masks for 11M images. The authors publish the trained model weights with the aim of providing a foundation model for a task they call promptable segmentation. This task constitutes a generalization of interactive segmentation. In order to prove the generality of their method, they perform evaluations on datasets containing rare objects. \cite{sam_hq} discusses an extension of SAM, called HQ-SAM, that has been fine tuned with the HQSeg-44k dataset, a dataset containing high quality human annotated segmentation masks. 
Interactive segmentation provides the benefit of generating ground truth annotations during test-time. In \cite{kontogianni2020continuous,shi2023self}, the clicks annotated by the authors are employed as labels for single pixels to further optimize the model whilst being in use. The authors of \cite{lin2023sequential} perform online optimization based on past annotations. \cite{hao2022rais} briefly mentions the usage of intermediate masks without going into further detail.

\section{Online Adaptation Methods} 
\label{sec:method}
Whenever we use an interactive segmentation model, we can assume the model to be applied to more than one image in the usage domain. In addition to this, interactive segmentation has the property of generating high quality masks for objects without any previously existing ground truth, while the employed system is currently in use. These masks are created to be used as ground truth in the future, and can thus be exploited to progressively adapt the system to the domain to which it is being applied. In addition to this, we obtain the ground truth for a single pixel after each click, allowing us to directly adapt to the current image.
In order to ameliorate the user experience when using interactive segmentation systems, we are going to explore various existing techniques for the purpose of adapting pretrained models in an online fashion. By this, we mean that the model does not require any form of fine-tuning before being used. 
We are first going to provide a quick summary of the problem of interactive segmentation. Afterwards, we are going to present the various possible adaptation methods that can be applied to the foundation models.  

\subsection{Interactive Segmentation}
Interactive Segmentation methods are usually conceptualized for the purpose of segmenting the surface of an object in an image by giving the system some form of user guidance \cite{ritm2022}. We are only looking at the case where the guidance is provided in the form of clicks/coordinates on the object's surface or the background, respectively. This constitutes a form of ground truth for single pixels which will help the system with the segmentation of the desired object. 
Interactive segmentation starts out with an image $\mathbf{x}_\text{Img} \in \mathbb{R}^{H \times W \times 3}$. Our goal is to create a segmentation mask $\mathbf{m} \in \{0,1\}^{H \times W}$ for a particular object in the image. 

For this, we train a network $\Phi_\text{IntSeg}$ to predict a segmentation mask from the clicked coordinates and the images. In our case, we are going to look at a scenario in which $\Phi_\text{IntSeg}$ is also given a preexisting, potentially faulty segmentation mask we want to improve. Whenever we have no such mask, we give the network a mask purely consisting of zeros. This is the case whenever the user has just made the first click.
The network $\Phi_\text{IntSeg}$ is going to be applied in an iterative fashion in order to progressively improve the mask with each interaction (i.e. with each click). Let $\tau$ be the index of the current interactive step. 
The user inspects the currently estimated mask $\mathbf{m}_{\tau - 1}$ and places a click on a region that isn't yet correctly labeled. These clicks take the form $(i_\tau, j_\tau, l_\tau)$ with $(i_\tau, j_\tau)$ being the coordinate on the image and $l_\tau$ being the label. This label either indicates the pixel as background ($l_\tau = -$) or foreground ($l_\tau = +$). In application scenarios, the label is usually indicated by using the right or left mouse button, respectively. 
Together with this new click, the notation $\mathbf{p}_{1:\tau} = \{\mathbf{p}_1, ..., \mathbf{p}_\tau \}$ represents all so far accumulated clicks. The interactive segmentation network then predicts a corrected mask $\mathbf{m}_\tau = \Phi_\text{IntSeg}(\mathbf{x}_\text{Img}, \mathbf{p}_{1:\tau}, \mathbf{m}_{\tau-1})$. 
These steps are repeated until the user judges the mask quality to be sufficient. 
Since this judgement is subjective, the mask may still contain incorrectly annotated areas.
We will call this resulting mask $\mathbf{m}^\text{Result}$. 

We are going to look at two interactive segmentation models: SAM \cite{kirillov2023segment} and HQ-SAM \cite{sam_hq}. We deem these models to be of elevated interest, because they are considered as \emph{foundation models} by the authors. This characterization specifically implies the ambition of creating a model that has been trained with such an enormous amount of training data, that it can be successfully applied to any arbitrary domain without having to be fine-tuned. 
The architectures of these two models can be divided into three parts: 
\begin{itemize}
    \item The \textbf{image encoder} is a vision transformer (see \cite{dosovitskiy2020image}) that receives the image and outputs feature maps describing the input image. 
    \item The \textbf{prompt encoder} encodes the clicks and the previous masks into an internal feature representation. 
    \item The \textbf{mask decoder} predicts the mask from the encoded prompts and the extracted image features.
\end{itemize}
This type of task division allows for an increased processing speed during interaction. The image only has to be processed once in order to extract the features. During the interaction, the generation of a predicted mask only requires the repeated execution of the prompt encoder and the mask decoder. As long as these two networks are sufficiently lightweight, the user can adjust the mask in real time. 
It should be noted that SAM has been designed for the more general task of promptable segmentation, which also involves bounding boxes or text as user guidance. In our experiments, however, we only look at click-based interactive segmentation. 
SAM has been trained on the dataset SA-1B, which contains 1.1B segmentation masks in 11M different images. This dataset has been published together with SAM and only contains automatically generated masks. 
The authors of HQ-SAM argue that purely training a model on automatically generated labels might lower the performance due to low-quality masks as the ground truth for the dataset. They propose a slight modification of the SAM architecture which they call HQ-SAM. In addition to the large pretraining of SAM, HQ-SAM is fine-tuned on a dataset called  HQSeg-44k, which contains high quality human-annotated masks.

\subsection{Click Adaptation (CA) and Click Mask (CM)} 
Right after the placement of a click in the image, we obtain the ground truth information for a single pixel. Over the course of $\tau$ interactions on said image, we accumulate progressively more annotated pixels $\mathbf{p}_{1: \tau}$. We can use these pixel as pseudo labels to adapt our model directly to the image (see \cite{kontogianni2020continuous,shi2023self}). 
This is done by creating a sparse mask which indicates that only some pixels are annotated by labeling the pixels without any label with the class $-1$:
\begin{equation}
\label{eq:sparsemask}
\mathbf{m}^\text{Sparse}_{\tau, i, j} = 
\begin{cases}
    1, &\text{ if } (i, j, +) \in \mathbf{p}_{1:\tau}\\ 
    0, &\text{ if } (i, j, -) \in \mathbf{p}_{1:\tau}\\
    -1, &\text{ otherwise } 
\end{cases}
\end{equation}

We can then employ this mask to directly compute and minimize the sparse binary cross-entropy 
\begin{equation}
\begin{split}
    \mathcal{L}_\text{Sparse}(\mathbf{m}^\text{Sparse}_\tau, \mathbf{m}_\tau) &= \frac{\sum_{i, j} 1_{\mathbf{m}^\text{Sparse}_{\tau, i, j} = 1} \mathcal{L}_\text{BCE}(\mathbf{m}^\text{Sparse}_{\tau, i, j}, \mathbf{m}_{\tau,i, j})}{\sum_{i, j} 1_{\mathbf{m}^\text{Sparse}_{\tau, i, j} = 1}} \\
    &+ \frac{\sum_{i, j} 1_{\mathbf{m}^\text{Sparse}_{\tau, i, j} = 0} \mathcal{L}_\text{BCE}(\mathbf{m}^\text{Sparse}_{\tau, i, j}, \mathbf{m}_{\tau, i, j})}{\sum_{x,y} 1_{\mathbf{m}^\text{Sparse}_{\tau, i, j} = 0}}
\end{split}
\end{equation}
with $\mathbf{m}_\tau$ being the currently estimated mask. 
We use this loss to partially optimize the network parameters.
This optimization poses a form of intentional overfitting to the current image. We will call this \textbf{click adaptation (CA)}. In our experiments (see \Cref{tab:sam_winter_85,tab:hqsam_winter_85}) we look at the options to either reverse this overfitting (marked with an \textbf{R}) after each image, or simply carry on without reversing (marked with a \textbf{C}). 

We also have the possibility of using the sparse mask to carry out an optimization after each image, in order to adapt the model to the images domain as a whole. We refer to this optimization as \textbf{click mask (CM)}. In \Cref{tab:sam_winter_85} and \Cref{tab:hqsam_winter_85} we use a \textbf{checkmark (\checkmark)} whenever this technique is employed. 

\begin{table*}[t]
    \centering
    \caption{The results on the WSESeg dataset for SAM. NoC means the $\text{NoC}_{20}@85$ metric and FR is the $\text{FR}_{20}@85$, describing the number of objects that could not be segmented after 20 clicks. For both metrics, a smaller value indicates a better performance. An explanation of the configurations can be found in \Cref{sec:method}: Click Adaptation (CA) can either be carried out with (R)esets or (C)ontinuously. The resulting mask (RM) can either be used (U)ntreated, pruned with (E)rosion or confidence thresholding (CT). We may also use all accumulated clicks as a click mask (CM). \label{tab:sam_winter_85} } 
    \begin{tabular}{|c|c|c|c|c|c|c|c|c|c|c|c|c|c|c|c|c|}
        \hline
        \multicolumn{6}{|c|}{Configuration} & \multicolumn{2}{|c|}{Bobsleigh} & \multicolumn{2}{|c|}{Curl. Stone}  & \multicolumn{2}{|c|}{Ski Helmet}   & \multicolumn{2}{|c|}{Snow Kite} & \multicolumn{2}{|c|}{Ski (Jump)} \\
        \hline
        \mcz{CA} & \mcz{RM} & \mcz{CM} & NoC & FR & NoC & FR & NoC & FR & NoC & FR & NoC & FR \\
        \hline
        \mcz{} & \mcz{} & \mcz{} & 3.379 &  3.39 &  3.586 &  9.56 &  9.050 &  33.04 &  5.972 &  23.96 &  16.050 &  74.50   \\ 
        \hline
        \mcz{R} & \mcz{E} & \mcz{$\checkmark$} & 3.502 &  3.06 &  3.617 &  9.36 &  8.173 &  26.17 &  5.996 &  23.96 &  15.323 &  \textbf{68.67} \\
        \mcz{R} & \mcz{CT} & \mcz{$\checkmark$} & 3.511 &  3.06 &  3.641 &  9.66 &  8.151 &  25.88 &  5.983 &  23.60 &  \textbf{15.229} &  68.88 \\
        \hline
        \mcz{R} & \mcz{} & \mcz{} & 3.415 &  3.06 &  3.481 &  8.75 &  \textbf{7.972} &  25.29 &  5.966 &  23.78 &  15.329 &  68.67 \\
        \mcz{C} & \mcz{} & \mcz{} & 3.450 &  3.06 &  3.452 &  8.65 &  8.044 &  \textbf{25.00} &  \textbf{5.926} &  \textbf{23.51} &  15.299 &  68.88 \\
        \mcz{} & \mcz{E} & \mcz{} & 3.440 &  3.06 &  3.579 &  9.26 &  8.792 &  31.14 &  5.987 &  24.05 &  15.763 &  72.89 \\ 
        \mcz{} & \mcz{CT} & \mcz{} & 3.445 &  2.90 &  3.574 &  9.16 &  8.542 &  29.24 &  5.981 &  23.87 &  15.707 &  72.29 \\
        \mcz{} & \mcz{} & \mcz{$\checkmark$} & \textbf{3.384} &  \textbf{2.74} &  3.585 &  9.36 &  8.692 &  30.12 &  6.004 &  24.13 &  15.677 &  71.69 \\ 
        \mcz{R} & \mcz{} & \mcz{$\checkmark$} & 3.461 &  3.06 &  \textbf{3.452} &  \textbf{8.55} &  7.985 &  25.15 &  5.955 &  23.69 &  15.337 &  68.67 \\ 
        \mcz{R} & \mcz{U} & \mcz{$\checkmark$} & 3.495 &  3.23 &  3.617 &  9.26 &  8.218 &  27.34 &  6.008 &  23.69 &  15.251 &  68.67 \\
        \hline

        \hline         
        \multicolumn{6}{|c|}{Curl. Broom} & \multicolumn{2}{|c|}{Ski Goggles} & \multicolumn{2}{|c|}{Ski (Misc)} & \multicolumn{2}{|c|}{Slalom Gate Poles} & \multicolumn{2}{|c|}{Snowboards} &  \multicolumn{2}{|c|}{\cellcolor[HTML]{DDDDFF}\textbf{Average}} \\
        \hline
         \mcd{NoC} & \mcd{FR} & NoC & FR & NoC & FR & NoC & FR & NoC & FR & NoC & FR \\
         \hline
         \mcd{13.855} &  \mcd{62.20} &  10.942 &  44.07 &  12.153 &  52.91 &  6.649 &  22.92 &  6.678 &  23.85 & 8.831 & 35.03 \\ 
         \hline
         \mcd{13.213} &  \mcd{57.16} &  10.441 &  39.90 &  11.589 &  49.08 &  6.349 &  20.12 &  \textbf{6.522} &  \textbf{22.62} & 8.472 & 32.01 \\
         \mcd{13.133} &  \mcd{57.01} &  10.501 &  40.90 &  11.554 &  48.59 &  6.402 &  20.21 &  6.648 &  24.00 & 8.475 & 32.18 \\
         \hline
         \mcd{\textbf{13.076}} &  \mcd{\textbf{55.49}} &  10.508 &  40.07 &  11.434 &  \textbf{47.42} &  \textbf{6.119} &  \textbf{18.67} &  6.686 &  23.85 & \textbf{8.399} & \textbf{31.50} \\
         \mcd{13.108} &  \mcd{56.25} &  10.558 &  40.40 &  11.720 &  49.75 &  6.346 &  20.50 &  6.605 &  23.23 & 8.451 & 31.92 \\
         \mcd{13.640} &  \mcd{60.37} &  10.741 &  42.57 &  11.854 &  50.75 &  6.565 &  22.15 &  6.734 &  24.15 & 8.710 & 34.04 \\
         \mcd{13.555} &  \mcd{59.60} &  10.618 &  41.74 &  11.927 &  51.75 &  6.640 &  22.15 &  6.694 &  23.85 & 8.668 & 33.65 \\
         \mcd{13.773} &  \mcd{60.82} &  10.920 &  43.91 &  11.880 &  51.08 &  6.742 &  22.82 &  6.669 &  23.85 & 8.732 & 34.05 \\ 
         \mcd{13.189} &  \mcd{56.25} &  \textbf{10.372} &  \textbf{39.57} &  \textbf{11.399} &  47.75 &  6.359 &  20.02 &  6.592 &  23.69 & 8.410 & 31.64 \\ 
         \mcd{13.392} &  \mcd{58.23} &  10.466 &  40.23 &  11.654 &  48.75 &  6.465 &  20.89 &  6.672 &  24.00 & 8.524 & 32.43 \\
         \hline
    \end{tabular}
    
\end{table*}

\begin{table*}[t]
    \centering
    \caption{The results on the WSESeg dataset for HQ-SAM. NoC means the $\text{NoC}_{20}@85$ metric and FR is the $\text{FR}_{20}@85$, describing the number of objects that could not be segmented after 20 clicks. For both metrics, a smaller value indicates a better performance. An explanation of the configurations can be found in \Cref{sec:method}: Click Adaptation (CA) can either be carried out with (R)esets or (C)ontinuously. The resulting mask (RM) can either be used (U)ntreated, pruned with (E)rosion or confidence thresholding (CT). We may also use all accumulated clicks as a click mask (CM). \label{tab:hqsam_winter_85} } 
    
    \begin{tabular}{|c|c|c|c|c|c|c|c|c|c|c|c|c|c|c|c|c|}
        \hline
        \multicolumn{6}{|c|}{Configuration} & \multicolumn{2}{|c|}{Bobsleigh} & \multicolumn{2}{|c|}{Curl. Stone}  & \multicolumn{2}{|c|}{Ski Helmet}   & \multicolumn{2}{|c|}{Snow Kite} & \multicolumn{2}{|c|}{Ski (Jump)} \\
        \hline
        \mcz{CA} & \mcz{RM} & \mcz{CM} & NoC & FR & NoC & FR & NoC & FR & NoC & FR & NoC & FR \\
        \hline
        \mcz{} & \mcz{} & \mcz{} & 8.961 &  26.61 &  10.897 &  43.34 &  18.785 &  91.23 &  8.335 &  34.69 &  19.189 &  94.78   \\ 
        \hline
        \mcz{R} & \mcz{E} & \mcz{$\checkmark$} & \textbf{5.219} &  \textbf{8.55} &  8.531 &  27.67 &  14.208 &  53.51 &  8.363 &  34.34 &  18.926 &  91.16 \\
        \mcz{R} & \mcz{CT} & \mcz{$\checkmark$} & 8.040 &  21.61 &  11.171 &  44.86 &  18.443 &  88.30 &  8.618 &  35.94 &  19.155 &  94.38 \\
        \hline
        \mcz{R} & \mcz{} & \mcz{} & 6.961 &  11.29 &  4.802 &  12.61 &  9.088 &  27.49 &  \textbf{7.738} &  \textbf{31.32} &  17.394 &  79.72 \\
        \mcz{C} & \mcz{} & \mcz{} & 7.276 &  14.03 &  \textbf{4.583} &  \textbf{12.31} &  \textbf{8.873} &  \textbf{26.46} &  8.118 &  33.27 &  16.578 &  74.70 \\
        \mcz{} & \mcz{E} & \mcz{} & 6.284 &  12.42 &  9.627 &  36.22 &  18.864 &  91.81 &  8.192 &  34.07 &  19.442 &  96.18 \\
        \mcz{} & \mcz{CT} & \mcz{} & 8.387 &  22.58 &  11.151 &  44.66 &  18.794 &  91.23 &  8.512 &  35.58 &  19.195 &  94.78 \\
        \mcz{} & \mcz{} & \mcz{$\checkmark$} & 17.248 &  80.81 &  15.260 &  68.67 &  19.067 &  92.98 &  9.594 &  40.20 &  19.038 &  93.78 \\ 
        \mcz{R} & \mcz{} & \mcz{$\checkmark$} & 13.229 &  44.19 &  7.411 &  20.55 &  9.418 &  28.65 &  9.447 &  39.22 &  \textbf{16.247} &  \textbf{73.09} \\ 
        \mcz{R} & \mcz{U} & \mcz{$\checkmark$} & 5.918 &  10.81 &  9.072 &  28.38 &  14.361 &  54.09 &  8.038 &  33.81 &  19.062 &  92.97 \\
        \hline

        \hline         
        \multicolumn{6}{|c|}{Curl. Broom} & \multicolumn{2}{|c|}{Ski Goggles} & \multicolumn{2}{|c|}{Ski (Misc)} & \multicolumn{2}{|c|}{Slalom Gate Poles} & \multicolumn{2}{|c|}{Snowboards} & \multicolumn{2}{|c|}{\cellcolor[HTML]{DDDDFF}\textbf{Average}} \\
        \hline
         \mcd{NoC} & \mcd{FR} & NoC & FR & NoC & FR & NoC & FR & NoC & FR & NoC & FR \\
         \hline
         \mcd{18.605} &  \mcd{90.70} &  17.424 &  80.80 &  17.504 &  84.36 &  13.480 &  60.54 &  11.215 &  49.38 & 14.440 & 65.64 \\ 
         \hline
         \mcd{18.460} &  \mcd{87.96} &  14.154 &  59.10 &  14.423 &  60.90 &  14.339 &  57.93 &  10.743 &  45.23 & 12.737 & 52.63 \\
         \mcd{18.957} &  \mcd{92.99} &  17.691 &  82.30 &  17.571 &  85.02 &  14.276 &  64.80 &  11.843 &  53.08 & 14.576 & 66.33 \\
         \hline
         \mcd{14.329} &  \mcd{60.37} &  \textbf{11.456} &  \textbf{39.40} &  \textbf{12.659} &  50.75 &  9.511 &  32.11 &  7.791 &  28.31 & 10.173 & 37.34 \\
         \mcd{\textbf{14.220}} &  \mcd{\textbf{59.91}} &  11.456 &  39.73 &  12.664 &  \textbf{50.58} &  9.952 &  34.43 &  \textbf{7.637} &  \textbf{27.38} & \textbf{10.136} & \textbf{37.28} \\
         \mcd{19.041} &  \mcd{93.45} &  17.678 &  82.64 &  17.539 &  85.02 &  15.103 &  67.89 &  11.251 &  49.38 & 14.302 & 64.91 \\ 
         \mcd{18.587} &  \mcd{90.55} &  17.372 &  80.63 &  17.567 &  84.86 &  13.446 &  60.06 &  11.257 &  49.85 & 14.427 & 65.48 \\
         \mcd{18.991} &  \mcd{93.60} &  18.526 &  87.15 &  18.268 &  89.35 &  15.484 &  71.76 &  12.720 &  58.15 & 16.420 & 77.64 \\ 
         \mcd{15.970} &  \mcd{71.49} &  12.137 &  43.24 &  15.068 &  65.89 &  \textbf{8.809} &  \textbf{30.27} &  8.914 &  33.69 & 11.665 & 45.03 \\ 
         \mcd{18.672} &  \mcd{89.79} &  16.634 &  72.12 &  16.128 &  74.88 &  16.479 &  75.73 &  10.865 &  46.31 & 13.523 & 57.89 \\
         \hline
    \end{tabular}
\end{table*}

\subsection{Using the Resulting Mask (RM) as Ground Truth}
The purpose of interactive segmentation is the creation of high quality object masks by a user. Thus we obtain a mask for a single object, once the user is done annotating. We can use the resulting mask as a pseudo label to adapt our model to the usage domain.
Since the quality of the mask depends on what the user regards as sufficient, there still may be erroneous areas on the mask. 
In order to allow the loss function to ignore these areas we tried two different methods of discarding potentially faulty mask pixels.  
The first approach is based on the assumption, that the erroneous areas of the resulting mask are most likely to be found close to the borders between foreground and background. 
We will therefore erode the background and foreground areas and ignore the eroded parts of the mask. 
We carry out an iterative form of erosion. Let $\mathbf{m}$ be a mask. Then we define $k$-fold iterative erosion to be 
\begin{align}
    \gamma^0(\mathbf{m}) &= \mathbf{m}, \\
    \gamma^{k+1}(\mathbf{m}) &= \gamma^k(\mathbf{m}) \ominus \begin{bmatrix}
        0 & 1 & 0 \\
        1 & 1 & 1 \\ 
        0 & 1 & 0
    \end{bmatrix}. 
\end{align}
This erosion operation is applied once to the background mask and once to the foreground mask. The results are then merged again into a new mask, where the eroded areas are set to the ignore label ($-1$). 
    
The second approach is to use confidence based pseudo-label filtering. The mask generated by the interactive segmentation model after the last user interaction can be seen as a probability map $\mathbf{m}^\text{Result, p}$. For each pixel, probabilities closer to $0$ indicate a higher likelihood of the pixel belonging to the background and probabilities closer to $1$ a higher likelihood of belonging to the foreground.
The value $0.5$ then corresponds exactly to the decision border between foreground and background. 
We define a confidence threshold $\delta \in (0, 0.5)$ which indicates how far the probability should be away from the decision border $0.5$ in order to be used as a label during optimization. Only pixels whose probability values have a distance of at least $\delta$ from $0.5$ are considered during loss computation (i.e. $|\mathbf{m}^\text{Result, p}_{i,j} - 0.5| \ge \delta$). 
In \Cref{tab:sam_winter_85,tab:hqsam_winter_85} the column \textbf{RM} indicates in what way we use the result mask during optimization. \textbf{E} stand for label filtering by erosion, \textbf{CT} stands for the application of a confidence threshold and \textbf{U} for using an unfiltered result mask.


\section{Experiments}

\subsection{Evaluation Details}
In order to carry out an automatic evaluation, we follow common practice (see \cite{ritm2022}) to simulate a plausible approximation to the behaviour of a human user. Note that the act of evaluating our system requires the availability of the ground truth for the images we evaluate on. Each click is generated in the following way: 
\begin{enumerate}
    \item Compare the predicted mask with the ground truth to obtain segmentation masks covering the false positive  and false negative areas.
    \item Compute $\mathfrak{D}_\textbf{FP}$ and $\mathfrak{D}_\textbf{FN}$, which are the distance transforms (see \cite{felzenszwalb2012distance}) of the false positive and false negative area respectively. 
    \item Find the coordinates $\mathbf{p}_\text{max, FP}$ and $\mathbf{p}_\text{max, FN}$ at which you can find the maximum values of the respective distance transforms. The new click will be placed at one of these two coordinates. 
    \item If $\mathfrak{D}_\textbf{FP}[\mathbf{p}_\text{max, FP}] > \mathfrak{D}_\textbf{FN}[\mathbf{p}_\text{max, FN}]$, place the new click at $\mathbf{p}_\text{max, FP}$ and give it a background ($-$) label.  Otherwise, place the new click at $\mathbf{p}_\text{max, FN}$ and give it a foreground ($+$) label.
\end{enumerate}

In order to ensure efficient processing by the model, we only optimize the mask decoder. In this way, the large encoder only has to be executed once.
We utilize the SAM and HQ-SAM architectures with a ViT-b backbone.  
We discovered appropriate values for hyperparameters in preliminary experiments: For adaptation, we use the Adam optimizer \cite{kingma2014adam} with a learning rate of $5 \cdot 10^{-8}$ for SAM and $10^{-6}$ for HQ-SAM.
We found $k=5$ to be the best performing number of iterations for iterative erosion, and $\delta=0.45$ to be the best performing confidence threshold. 

Since our experiments are going to measure the usability of SAM and HQ-SAM as interactive segmentation systems, we are going to measure the $\text{NoC}_{20}@85$ and $\text{FR}_{20}@85$ metrics, which are computed as follows: 
Let $\mathbf{m}_\text{GT}$ and $\mathbf{m}_\tau$ be the ground truth mask and the currently estimated mask, respectively. We asses the mask quality by $\text{IoU}(\mathbf{m}_\text{GT}, \mathbf{m}_\tau) = \frac{\mathbf{m}_\text{GT} \cap  \mathbf{m}_\tau}{\mathbf{m}_\text{GT} \cup \mathbf{m}_\tau}$. 
For both metrics, we have a certain IoU threshold (in our case 85). In order for a mask to be regarded as sufficiently well annotated, the overlap with a preexisting ground truth has to exceed this IoU threshold. In the domain of interactive segmentation, $\text{NoC}_{20}@85$ measures the number of click interactions necessary to reach this threshold. 
This number of interactions is capped to 20 foreground / background clicks. 
This is also the same concept used for the failure rate $\text{FR}_{20}@85$: If the predicted mask does not reach an IoU of at least 85 with the ground truth after 20 clicks, we consider the system to have failed annotating this object. The $\text{FR}_{20}@85$ measures this failure rate. Out of the two metrics we regard the failure rate as slightly more important, since it captures a form of inapplicability of the model.

\subsection{Results}
We are first going to look at the performance of our method when being applied to the SAM model. 
For this, we use the $\text{NoC}_{20}@85$ and $\text{FR}_{20}@85$ metric as a measure of performance. 
The results for this setting can be seen in \Cref{tab:sam_winter_85}. 
On both metrics the best average result is achieved by click adaptation with resets, without using any form of dense mask. Here the FR is reduced from 35.02 to 31.50 while the NoC is reduced from 8.831 to 8.399. 
This is however not the case for all classes: The lowest FR and NoC on the snowboards class is achieved by a mixture of click adaptation with resets and using the eroded result masks, which incurs a reduction of the FR to 22.62 percentage points and the NoC to 6.522 clicks.
On the bobsleigh class we reach the lowest FR only using the click mask. 
Pruning the result mask incurs an improvement with both strategies. Using erosion reduces the failure rate by 0.99 percentage points while using confidence thresholding reduces the FR by 1.38 percentage points. If we combine these methods with the click mask and click adaptation with resets, the FR is reduced even further by 3.02 percentage points for erosion and 2.85 percentage points for confidence thresholding. 

We also test our method on a second type of model: HQ-SAM, which is an slightly altered extension of SAM that has been fine-tuned on the HQSeg-44K dataset. The authors of \cite{sam_hq}, which proposed HQ-SAM, criticized the lack of fine grained quality in the masks predicted by SAM. The results obtained by HQ-SAM can be seen in \Cref{tab:hqsam_winter_85}. 
The HQSeg-44k dataset provides high-quality human-made annotations, in contrast to SA-1B, in which the annotations have been automatically generated. While this fine-tuning results in a general improvement of mask quality, it also specializes the architecture to a vastly smaller variety of data. This turns out to be detrimental when applying the resulting model to new domains, such as winter sports equipment. This can especially be seen by the drastically increased average  failure rate of 65.64 for HQ-SAM vs. 35.03 for SAM. 
Here, click adaptation without resets incurs the biggest improvement, reducing the failure rate from 65.64 to 37.28 on and the NoC from 14.440 to 10.136. For click adaptation with resets, the failure rate is reduced to 37.34. 
On the class of bobsleighs we can see a deviation from this trend, where the combination of click adaptation with resets, using the click mask and the eroded result mask reduces the failure rate from 26.61 to 8.55 and the NoC from 8.961 to 5.219. We can observe a similar behavior on the slalom gate pole class, where the best results are achieved by a combination of the click mask with click adaptation with resets. While there is no single best method for all types of objects, we can reach notable improvements employing adaptation methods on average.

\section{Conclusion}
In this paper we presented a new dataset containing instance segmentation masks on winter sports equipment of ten different classes. 
With the publication of this dataset we aim at supporting research in the direction of applying computer vision for sports analysis and  human body part segmentation. 
Since our dataset contains rare classes, it provides a viable way of testing foundation models for interactive segmentation. 
We test the performance of two foundation models, SAM and HQ-SAM. In order to look at ways of improving the user experience regarding such models, we also test various methods to adapt said foundation models to the winter sports domain during usage. 
Despite being conceptualized as foundation models, their performance can be improved using test-time adaptation methods.


\bibliography{references}{}
\bibliographystyle{unsrt}

\end{document}